\documentclass{article}

\usepackage{PRIMEarxiv}

\usepackage[utf8]{inputenc} 
\usepackage[T1]{fontenc}    
\usepackage{hyperref}       
\usepackage{url}            
\usepackage{booktabs}       
\usepackage{amsfonts}       
\usepackage{nicefrac}       
\usepackage{microtype}      
\usepackage{lipsum}
\usepackage{fancyhdr}       
\usepackage{graphicx}       
\usepackage{subfigure}
\usepackage{multicol}
\usepackage{authblk}

\makeatletter
\newenvironment{tablehere}
  {\def\@captype{table}}
  {}

\newenvironment{figurehere}
  {\def\@captype{figure}}
  {}
\makeatother

\graphicspath{{media/}}     

\title{TE-YOLOF: Tiny and efficient YOLOF for blood cell detection
}

\author[a]{Fanxin Xu}
\author[b]{Xiangkui Li}
\author[a]{Hang Yang}
\author[a]{Yali Wang}
\author[a,†]{Wei Xiang} 

\affil[a]{College of Electronic and Information, Southwest Minzu University}
\affil[b]{West China Biomedical Big Data Center}
\affil[ ]{\{xufanxin, yanghang, wangyali\}@stu.swun.edu.cn, lixiangkui@wchscu.cn, 21500068@swun.edu.cn}

\begin{document}
\maketitle
\footnotetext{† Corresponding author.}

\begin{abstract}
Blood cell detection in microscopic images is an essential branch of medical image processing research. Since disease detection based on manual checking of blood cells is time-consuming and full of errors, testing of blood cells using object detectors with Deep Convolutional Neural Network can be regarded as a feasible solution. In this work, an object detector based on YOLOF has been proposed to detect blood cell objects such as red blood cells, white blood cells and platelets. This object detector is called TE-YOLOF, Tiny and Efficient YOLOF, and it is a One-Stage detector using dilated encoder to extract information from single-level feature maps. For increasing efficiency and flexibility, the EfficientNet Convolutional Neural Network is utilized as the backbone for the proposed object detector. Furthermore, the Depthwise Separable Convolution is applied to enhance the performance and minimize the parameters of the network. In addition, the Mish activation function is employed to increase the precision. Extensive experiments on the BCCD dataset prove the effectiveness of the proposed model, which is more efficient than other existing studies for blood cell detection.
\end{abstract}

\hspace*{\fill}

\begin{multicols}{2}
\section{Introduction}

The analysis of blood cell in microscopic images plays a vital role in disease recognition field by identifying the different cellular objects. In blood cell field, there are three important components in blood: White Blood Cells(WBC), Red Blood Cells(RBC), and Platelets \cite{Atkins2017RamanSO}. The proportion and number of these blood cells are seriously affect the doctor's judgment of the illness \cite{BurgaraEstrella2020AtomicFM, LpezCanizales2021NanoscaleCO}. Finding an automated algorithm based on deep convolutional neural network to detect blood cells accurately and efficiently can improve the effectiveness of the medical system \cite{Tiwari2018DetectionOS}. 

Object detection which is the study of finding the coordinates of objects in an image and classifying objects is widely used in computer vision tasks, such as machine vision, pedestrian identification, abnormal detection and so on. Fast, accurate algorithms for object detection would allow computers to instead of part of manual checking instances and unlock the potential of humans for the better purpose.

Current object detectors are either based on two-stage or on one-stage mechanism \cite{Wu2020RecentAI}. In two-stage field, most networks are based on the R-CNN framework which generates a sparse set of candidate object locations in the first stage and classifies each candidate location as the foreground classes or background in the second stage \cite{Girshick2014RichFH, Girshick2015FastR, Ren2015FasterRT, He2020MaskR}. Through a sequence of advances, such two-stage models reach highest accuracy with slow operation. While one-stage detectors, such as YOLO(You Only Look Once) and SSD(Single Shot MultiBox Detector), reframe object detection as a single regression problem by learning the class probabilities and bounding box coordinates in a freeway \cite{Redmon2016YouOL, Redmon2017YOLO9000BF, Redmon2018YOLOv3AI, Liu2016SSDSS}. With the help of focal loss, RetinaNet\cite{Lin2020} which belong to one-stage detectors outperforms the alternative high-accuracy to two-stage detectors. Cascading dilated convolutions is utilized in YOLOF(You Only Look One-level Feature)\cite{chen2021you} to obtain the same effect on dense small sample detection by feature pyramid networks(FPN)\cite{Lin2017FeaturePN}.

Inspired by \cite{Shakarami2021}, in this research, we propose a new one-stage detector that provide high accuracy besides high efficiency to solve the problem of the blood cell detection with low accuracy. In a nutshell, the contributions of this research are divided into the following items:

\begin{itemize}

\item Applying YOLOF to the blood cell detection field for the first time and using the EfficientNet CNN as the backbone to increase efficiency and flexibility.

\item Depthwise Separable Convolution module is applied to enhance the performance and minimize the parameters in the decoder.

\item Mish activation function is proved to be the relatively optimal method to increase the precision compared to other functions, such as Swish, ReLU, MetaAconC and so on.

\item Extensive experiments on BCCD dataset indicates the importance of each component. Moreover, we conduct comparisons with YOLOv3, Deformable DETR. We can achieve comparable results with a higher mAP.

\end{itemize}

\section{Background and literature review}

Before the advance of one-stage detectors, region proposals are used to locate and classify objects in images instead of sliding windows in two-stage mechanism\cite{Uijlings2013SelectiveSF, Girshick2014RichFH}. As Faster R-CNN came up, region proposals can be automatic generated by Region Proposal Network(RPN)\cite{Ren2015FasterRT}. So far, object detection is unified into a pure neural network framework without any hand-designed parts to extract features and predict the object. These complex pipelines can help two-stage detectors to achieve high accuracy at the expense of speed. 

In one-stage field, Multiple versions of YOLO series have been published to reframe object detection as a single regression problem \cite{Redmon2016YouOL, Redmon2017YOLO9000BF, Redmon2018YOLOv3AI}, they forward an image only once to predict where and what objects are present. Feature Pyramid Network(FPN) have beed used in YOLOv3 to realize object detection of different sizes. These methods are much faster than the two-stage detectors with comparable accuracy. With the help of focal loss that enable us to train a high accuracy by addressing the class imbalance problem\cite{Lin2020}, one-stage detectors are able to achieve an excellent balance of precision and speed. While in\cite{chen2021you}, YOLOF use dilated encoder to obtain the feature maps that has the comparable effect as FPN, bridging the performance gap between detectors with FPN and others without FPN in small object detection. In the following, detailed approaches that we utilized in blood cell detection have been described with extensive experiences.

In a blood cell image, the distribution of WBC and RBC are relatively unbalanced. The distribution of RBC is relatively dense with different sizes, while the distribution of WBC and platelet are relatively sparse. In \cite{Zhang2019}, YOLOv3 has been utilized to detect discrete RBC and WBC for the purpose of counting RBC and WBC in images. Image density estimation algorithm \cite{Lempitsky2010LearningTC} has been used to the counting tasks of RBC. In \cite{Zhang2019}, the maximum mAP is achieved as 88.26\% by the method of YOLOv3 with approximately 62 million parameters. While in \cite{Shakarami2021}, FED detector update the mAP to 88.33\% with approximately 14 million parameters. With the effect of Swish and DIoU, mAP is improved to 89.86\% . 

Depthwise separable convolutional network is proposed to reduce computation and model size. It has been used in neural network design by MobileNet\cite{Howard2017MobileNetsEC} and has become more popular since its inclusion in the TensorFlow framework. In the method proposed in this paper, depthwise separable convolution is presented for reducing the parameters and improve the effect.

Compound scaling method is used to balance network width, depth, and resolution according to a certain ratio. Before this method came up, increasing the depth of the network is the most common way to increase performance of the network\cite{He2016DeepRL, He2016IdentityMI}. Additionally, although the input image with higher resolution is helpful for feature extraction and better performance\cite{Howard2017MobileNetsEC, Sandler2018MobileNetV2IR}, it increases the number of parameters. EfficientNet use compound scaling method to achieve state-of-the-art accuracy with an order of magnitude fewer parameters and FLOPs on both ImageNet and other transfer learning datasets\cite{Tan2019EfficientNetRM}. Therefore, in this study, EfficientNet is applied to this research to increase efficiency.

\begin{figure*}[t]
    \centering
    \includegraphics[width=1\textwidth]{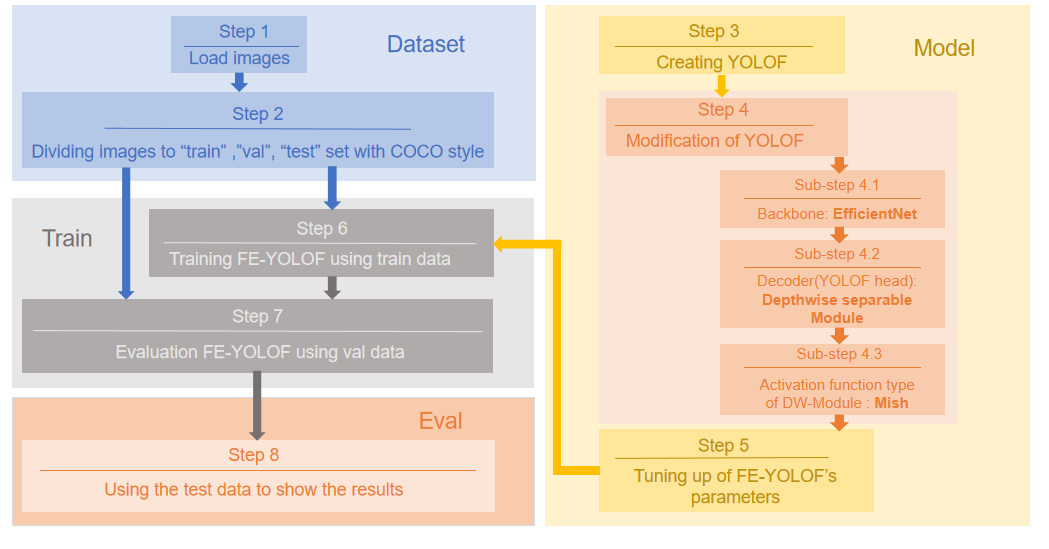}
    \caption{Flow chart of the proposed detector}
    \label{fig:flow_chart}
\end{figure*}

\section{Material and methods}

The goal of object detection is finding the coordinates of objects in an image and also classification of its category. In this research, the proposed detector regards object detection as a regression problem, which is equivalent to the methodology of the YOLO detector. Also the proposed detector that only use the last output feature of the backbone can converges faster and achieves promising performance, compared to YOLOv3 which use FPN to enhance performance. This proposed detector is modified by YOLOF which is detailed described in \cite{chen2021you}.

The flow chart of the proposed detector has been shown in Figure \ref{fig:flow_chart}. The entire operation of the proposed model includes 8 steps. The details of these steps are as follows.
\begin{itemize}
    \item Step 1: The images with data augumentation of BCCD dataset are loaded in a dataset.
    \item Step 2: The loaded images are divided into train, val and test set with annotation format of COCO style.
    \item Step 3: In this step, creating the YOLOF model in order to create the proposed model subsequently.
    \item Step 4: Overall architecture modification process of this model is detailed in this step. These modifications involve the backbone of the model, the use of the decoder module in the detection head and the selection of the activation function.
    
    \begin{itemize}
        \item Sub-step 4.1: This modification is mainly to change the backbone of the model. EfficientNet is applied to replace the original backbone of ResNet-50.
        \item Sub-step 4.2: Second major modification is using depthwise separable module to replace the normal convolution module in decoder layer of detector head.
        \item Sub-step 4.3: Another modification of the detector head is choosing the Mish activation function instead of ReLU with exhaustive experiment. The overall architecture of TE-YOLOF is depicted detailed in Figure \ref{fig:architecture}.
    \end{itemize}
    
    \item Step 5: Tuning up the parameters of the model. The parameters of the backbone use the parameters after ImageNet pre-training. Parameter settings in the training phase are explained with detailed description in sub-section 4.1 and other parameters, which are not mentioned, are tuned up similar to parameters of YOLOF\cite{chen2021you}.
    
    \item Step 6: The proposed detector, TE-YOLOF, is trained using the training data. Additional information about the training phase is detailed described in sub-section 4.2. 
    
    \item Step 7: The proposed model after the phase of training is evaluated in the val set. Mean Average Precision(mAP) is the mainly used indicator. In addition, the efficiency of the proposed detector is evaluated by the number of parameters and Giga Floating-point Operations Per Second(GFLOPs). These evaluations with exhaustive experiment have been detailed explained in sub-section 4.3.
    
    \item Step 8: The visualization results of the proposed detector have been shown in this step with using test data set. Detailed results are displayed in sub-section 4.3.
\end{itemize}

\subsection{TE-YOLOF}

\begin{figure*}[t]
    \centering
    \includegraphics[width=1\textwidth]{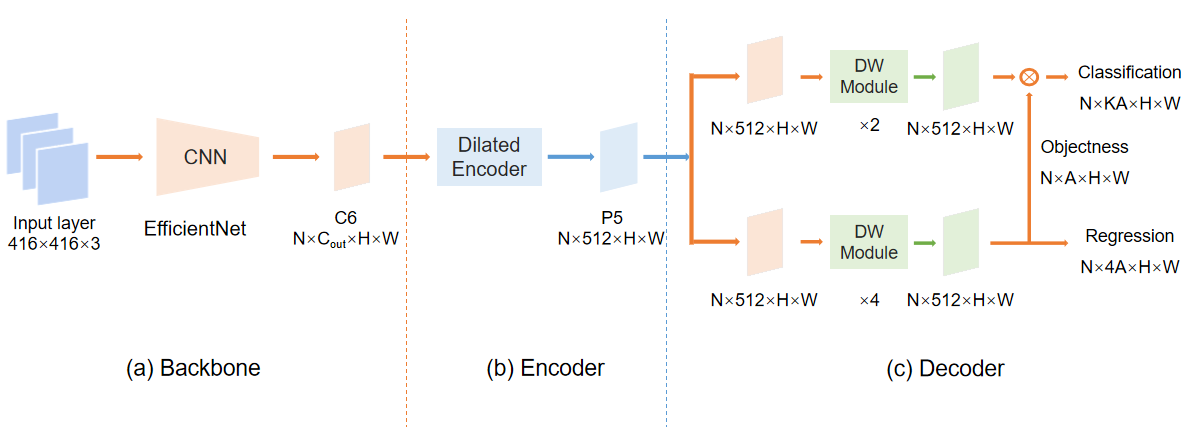}
    \caption{The sketch of TE-YOLOF, which consists of three main components: the backbone, the encoder, and the decoder. In the figure, input layer with 416 × 416 × 3 represents the size of each picture entered. 'C6' represents the output feature of the backbone with downsample rate of 32. ’$C_{out}$’ means the number of channels of the feature. The number of channels in the encoder and the decoder is seted as 512. H × W represents the height and width of feature maps.}
    \label{fig:architecture}
\end{figure*}

Based on the solutions above, we propose a lightweight framework as TE-YOLOF. The sketch of TE-YOLOF is shown in Figure \ref{fig:architecture}. The proposed framework consists of three parts: the backbone, the encoder, and the decoder. In this section, a brief introduction of the main components which we used in the proposed detector is given as follows.

\paragraph{Backbone.}
EfficientNet series are adopted as our backbone in all models. For the purpose of efficiency and flexibility, EfficientNet-B0 to EfficientNet-B3 are chose to analysis with the balance of precision and parameters. All models are pre-trained on ImageNet. 
The output of the backbone is the C6 feature map which has the default 1280 channels with different width enhancement factor in different backbones, and with a downsample rate of 32.

\begin{figurehere}
    \centering
    \includegraphics[width=0.48\textwidth]{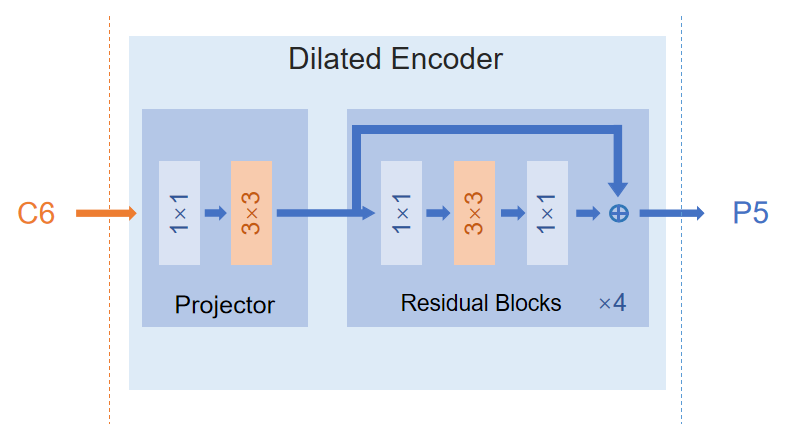}
    \caption{The architecture of Dilated Encoder. The Projector include two convolution layers(one 1 × 1 and one 3 × 3 convolution). 1 × 1 and 3 × 3 denotes 1 × 1 and 3 × 3 convolution layers. ×4 means four successive residual blocks.'C6' represents the output feature of the backbone. 'P5' represents the output feature of the Dilated Encoder.}
    \label{fig:dilated_encoder}
\end{figurehere}

\paragraph{Encoder.}
The specific architecture of encoder is shown in Figure \ref{fig:dilated_encoder}. The Projector is added after the backbone with two projection layers (one 1 × 1 and one 3 × 3 convolution), resulting in a feature map with 512 channels. Then using residual blocks, which consist of three main components: the first 1 × 1 convolution is applied to channel reduction with a reduction rate of 4, then a 3 × 3 convolution is used to enlarge the receptive field with different dilation factor in different block, a 1 × 1 convolution is employed to recover the number of channels at last.

\begin{figurehere}
    \centering
    \includegraphics[width=0.48\textwidth]{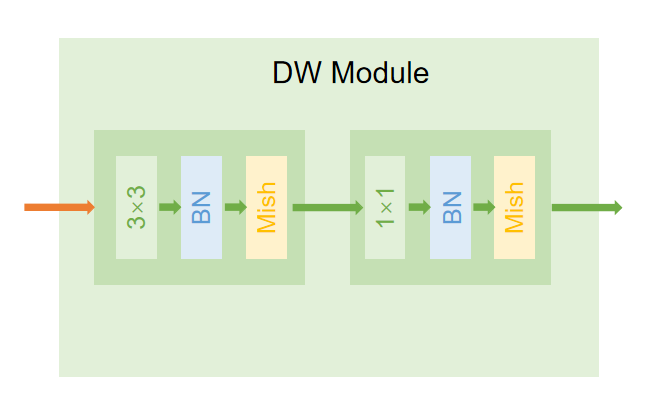}
    \caption{The architecture of Depthwise Separable Convolution Module. 3 × 3 denotes 3 × 3 convolution layer by channel-wise. 1 × 1 denotes 1 × 1 convolution layer. BN represents the batch normalization layer. Mish represents the activation function of Mish we used.}
    \label{fig:dw_module}
\end{figurehere}

\paragraph{Decoder.}
The decoder, which consists of two parallel task-specific heads: the classification head and the regression head, is shown in Figure \ref{fig:architecture}. There are four Depthwise Separable Convolution modules on the regression head while only have two on the classification head. The architecture of Depthwise Separable Convolution Module is shown in Figure \ref{fig:dw_module}. Each convolution layer followed by batch normalization layer and Mish layer in the module. We follow Autoassign \cite{Zhu2020AutoAssignDL} and use objectness prediction for each anchor on the regression head to prove whether the anchor containing the object. The final predictions of the classification scores are resulted by multiplying the classfication output with the objectness prediction.

\begin{figurehere}
    \centering
    \includegraphics[scale=0.4]{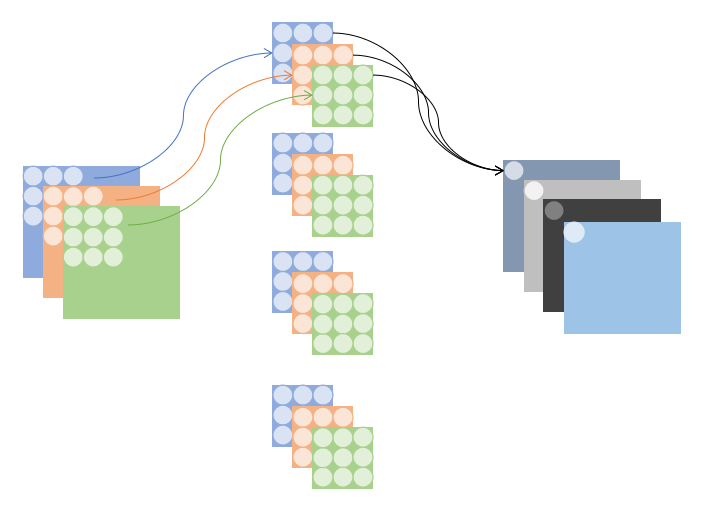}
    \caption{Standard Convolution}
    \label{fig:standard_convolution}
\end{figurehere}

\paragraph{Other Details.}
Non-Maximum Suppression is utilized in this object detection algorithm to be sure that the detector detects each object only once. The detected boxes which overlap the box with the highest score exceed a threshold would be removed. Focal loss which is presented to address the unbalance between positive and negative samples problem is used as the classification loss in this model. GIoU \cite{Rezatofighi2019GeneralizedIO} is utilized to solve the regression loss.

\subsection{EfficientNet with compound scaling method}
Compound scaling is an effective method to achieve higher performance by uniformly scaling network width, depth and resolution in a principled way. Although only increasing one of the factors can improve the accuracy, the network become too large with enormous amount of parameters and operations. A compound coefficient $\phi$ is used to balance different factors in a restricted condition and reach higher precision. The detailed accomplishment is in Equation (\ref{eq:compound}). 

\begin{eqnarray} \label{eq:compound}
    depth: d&=&\alpha^{\phi} \nonumber \\
    width: w&=&\beta^{\phi} \nonumber \\
    resolution: r&=&\gamma^{\phi} \\
    &s.t.&\alpha \cdot \beta^2 \cdot \gamma^2 \approx 2 \nonumber \\
    &&\alpha \ge 1, \beta \ge 1, \gamma \ge 1 \nonumber
\end{eqnarray}

Intuitively, $\phi$ is a user-specified parameter that controls how many more resources are available for model scaling, while $\alpha$, $\beta$, $\gamma$ determine the expansion of network width, depth, and resolution respectively. The best values for EfficientNet-B0 are $\alpha = 1.2, \beta = 1.1, \gamma = 1.15$. EfficientNet-B1 to B7 are obtained with different $\phi$ under Equation (\ref{eq:compound}).

In this study, for efficiency and flexibility, EfficientNet-B0 to B3 are chosen to analysis as backbone with 5.3, 7.8, 9.2, 12 million parameters respectively. On the other hand, different backbones are utilized to prove the effectiveness of the model between accuracy and resource.

\begin{figurehere}
    \centering
    \includegraphics[scale=0.33]{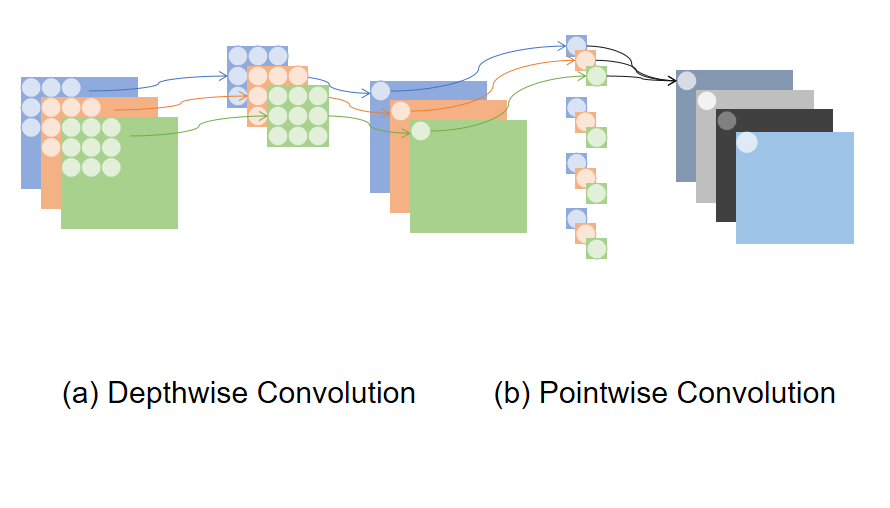}
    \caption{Depthwise Separable Convolution}
    \label{fig:depthwise_separable_convolution}
\end{figurehere}

\subsection{Depthwise Separable Convolution}

Depthwise Separable Convolution is an effective method to minimize the parameters by factorizing a standard convolution into a depthwise convolution and a 1 × 1 convolution called a pointwise convolution \cite{Howard2017MobileNetsEC, Li2019YOLOv3LiteAL}. For the proposed detector, the depthwise convolution applies a single filter to each input channel and the pointwise convolution applies a 1 × 1 convolution to combine the outputs from the depthwise convolution. The specific architecture is shown in Figure \ref{fig:dw_module}. The batch normalization layer and Mish activation function is followed after the convolution layer. 

For more details, the standard convolution is presented in Figure \ref{fig:standard_convolution} and the corresponding depth separable convolution is shown in Figure \ref{fig:depthwise_separable_convolution}. Suppose the input channel is $N$, the output channel is $M$, and the size of the convolution kernel is $N_k \times N_k$. According to the Figure \ref{fig:standard_convolution}, the standard convolution kernels is $N \times N_k \times N_k \times M$. As a reference in Figure \ref{fig:depthwise_separable_convolution}, the depthwise convolution kernels is $N \times N_k \times N_k$, while the pointwise convolution kernels is $ N \times 1 \times 1 \times M$. Compared with standard convolution, depthwise separable convolution uses 8 to 9 times less parameters.

\subsection{Mish activation function}

\begin{figure*}[t]
    \subfigure[Activation Method]
    {
        \begin{minipage}{8cm}
            \centering
            \includegraphics[scale=0.48]{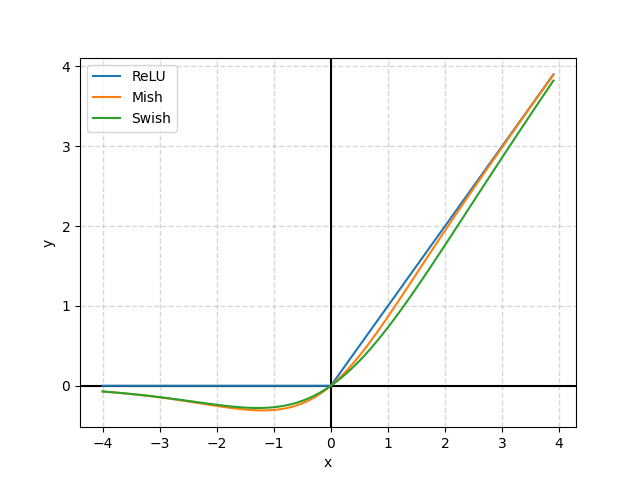}
            \label{fig:sub_activation}
        \end{minipage}
    }
    \subfigure[Derivate Activation Method]
    {
        \begin{minipage}{8cm}
            \centering
            \includegraphics[scale=0.48]{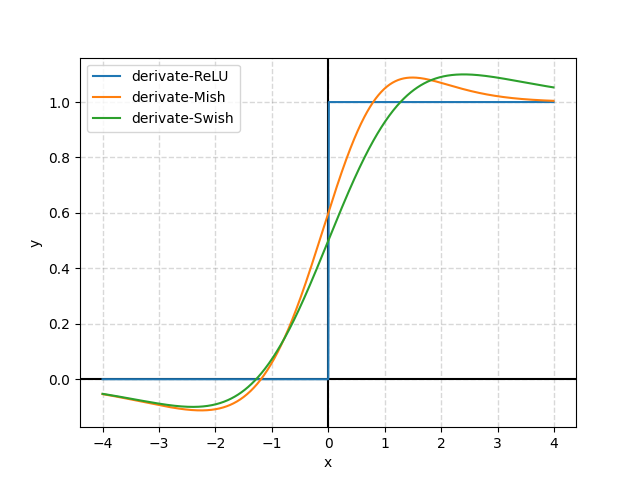}
            \label{fig:sub_derivate_activation}
        \end{minipage}
    }
    \caption{(a) Graph of ReLU, Mish and Swish activation functions. Mish and Swish are closely related with both having a distinctive negative concavity. (b) Graph of the derivatives of ReLU, Mish and Swish activation functions. }
    \label{fig:activation}
\end{figure*}

Mish is a smooth, continuous, self regularized, non-monotonic activation function mathematically defined as:
\begin{eqnarray} 
    \label{eq:mish}
    f(x)    &=& x \tanh (softplus(x)) \nonumber\\
            &=& x \tanh{(\ln(1+e^x))}
\end{eqnarray}
According to the Equation (\ref{eq:mish}), Mish uses the Self-Gating property where the non-modulated input is multiplied with the output of a non-linear function of the input. While the Swish is defined as $f(x) = x \cdot (1 + \exp{(-\beta x)})^{-1}$. Figure \ref{fig:activation} shows the graph of Swish for $\beta=1$ and Mish versus ReLU \cite{Misra2020MishAS, Krizhevsky2012ImageNetCW, Ramachandran2018SearchingFA}. As illustrated in the Figure \ref{fig:sub_activation}, Mish, similar to Swish, is bounded below and unbounded above with a range of $[\approx -0.31, \infty)$. As the Figure \ref{fig:sub_derivate_activation} shows that the Mish function converges to one faster than Swish in the positive value.

\begin{tablehere}
    
    \centering
    \begin{tabular}{p{4cm}p{3cm}}
        \toprule
        Parameter     & Type     \\
        \midrule
        Programming Language    & Python 3.7    \\
        Library and wrapper     & Pytorch   \\
        GPU                     & GeForce 1080Ti     \\
        RAM                     & 11 G \\
        Training algorithm      & SGD \\
        Evaluation metrics      & MS COCO detection evaluation metrics \\
        \bottomrule
        \\
    \end{tabular}
    \caption{Implementing tools}
    \label{tab:implementing}
\end{tablehere}

The benefits of Mish are given as bellows.
\begin{itemize}
    \item Mish eliminated the Dying ReLU phenomenon by design the preconditions to preserve a small amount of negative information.
    \item Mish is bounded below and avoids saturation by near-zero gradients. It is unbounded above so that the outputs do not saturate to the maximum value.
    \item Compared to ReLU and Swish, Mish is continuously differentiable and plays a role in better gradient flow.
\end{itemize}

\section{Results}
In this section, we focus on the implementation details and experiments analysis. We evaluate our detector on the BCCD dataset and conduct comparisons with Deformable DETR and YOLOv3. Then, we provide a detailed ablation study of each component's design with quantitative results and analysis. The details are as follows.

\begin{tablehere}
    
    \centering
    \begin{tabular}{p{2cm}lp{2cm}l}
        \toprule
        Data division & Type & Number of objects per category & Total \\
        \midrule
                & RBC       & 2938  &           \\
        Train   & WBC       & 263   &   3450    \\
                & Platelets & 249   &           \\
        \hline
                & RBC       & 819   &           \\
        Val     & WBC       & 72    &   967     \\
                & Platelets & 76    &           \\
        \hline
                & RBC       & 398   &           \\
        Test    & WBC       & 37    &   471     \\
                & Platelets & 36    &           \\
        \bottomrule
        \\
    \end{tabular}
    \caption{The division of the BCCD dataset}
    \label{tab:bccd_dataset}
\end{tablehere}

\paragraph{Implementation Details.}
\label{par:implementation}
In this study, Python and Pytorch, an Nvidia GeForce 1080Ti have been utilized for implementing the proposed detector. The implementing tools is 
detailed introducing in Table \ref{tab:implementing}. The proposed model is trained with SGD over 1 GPU with 4 images per mini-batch. The ablation study is based on the '1x' schedule setting with initial learning rate of 0.015. To stabilize the training at the beginning, we set the number of warmup iterations with 1500. NMS with a threshold of 0.6 is utilized to post-process the results.

\paragraph{Dataset.}
The BCCD dataset, which includes 364 images that each of them has dimentions of $640 \times 480 \times 3$ and contains different proportions of Platelets, WBC and RBC,has been used to evaluate the proposed detector. All the images are resize to $416 \times 416 \times 3$. The division protocol of this dataset has divided images into training, val and test sets with a ratio of 7:2:1, the detailed information of objects per categories is shown in Table \ref{tab:bccd_dataset}. In addition, horizontal flip, vertical flip, randomly crop between 0 and 15 percent of the image, random brigthness adjustment, random exposure adjustment, are applied to augmentation. All cells are annotated in COCO format.

\subsection{Comparison with previous works}

\begin{table*}[]
    \centering
    \begin{tabular}{cccccccccc}
        \toprule
        Model & Epochs & \#params & GFLOPS & $AP$ & $AP_{50}$ & $AP_{75}$ & $AP_S$ & $AP_M$ & $AP_L$ \\
        \midrule
        Deformable DETR & 50 & 40M & 173G & \textbf{60.0} & 87.7 & \textbf{71.9} & \textbf{44.5} & \textbf{61.3} & 47.1 \\
        YOLO v3 & 38 & 62M & 194G & 45.9 & 86.7 & 45.3 & 22.6 & 50.4 & 38.8 \\
        YOLOF & 38 & 42M & 98G & 57.5 & 89.0 & 65.7 & 43.8 & 57.9 & 47.3 \\
        YOLOF$+$ & 12 & 42M & 98G & 53.1 & 87.6 & 56.2 & 36.3 & 56.3 & 45.4 \\
        \hline
        TE-YOLOF-B0 & 12 & 9.94M & 6.21G & 55.6 & 89.4 & 59.8 & 35.1 & 58.2 & \textbf{48.1} \\
        TE-YOLOF-B1 & 12 & 12.45M & 6.32G & 57.8 & 90.1 & 64.5 & 38.4 & 62.2 & 46.1 \\
        TE-YOLOF-B2 & 12 & 13.7M & 6.4G & 56.6 & 90.1 & 60.4 & 37.1 & 59.1 & 47.8 \\
        TE-YOLOF-B3 & 12 & 16.76M & 6.6G & 58.4 & \textbf{90.6} & 66.0 & 40.4 & 59.6 & 47.5 \\
        \bottomrule
        \\
    \end{tabular}
    \caption{Comparison with Deformable DETR and YOLO v3 on the BCCD validation set. The top section shows the results of various models. The original setting of Deformable DETR \cite{Zhu2021DeformableDD} is applied to train and test. The setting of YOLO v3 is same to the method in \cite{Zhang2019} with batchsize of 8. YOLOF utilize the setting of the method in \cite{Shakarami2021} with AdamW \cite{Loshchilov2019DecoupledWD}. The $+$ represents that YOLOF use the same setting of TE-YOLOF shown in \textbf{\nameref{par:implementation}}. In the bottom section, for those models marked with suffix of 'B(N)', they adopt EfficientNet-B(N) as backbone by default.}
    \label{tab:comparison_previous}
\end{table*}

\begin{table*}[t]
    \centering
    \begin{tabular}{cccccc}
        \toprule
        Model           & \#params  & Platelets & RBC & WBC & mAP \\
        \midrule
        FED             & 13.68M    & 90.3  & 80.4  & 98.9  & 89.9 \\
        TE-YOLOF-B0     & 9.94M     & 91.1  & 82.8  & 98.2  & 90.7  \\
        \hline
        TE-YOLOF-B0*    & 9.94M     & 89.4  & 84.6  & 98.7  & 90.9  \\
        TE-YOLOF-B3*    & 16.76M    & 89.8  & 87.3  & 98.7  & 91.9  \\
        \bottomrule
        \\
    \end{tabular}
    \caption{Comparison with FED on the BCCD validation set. Platelets, RBC and WBC represents the AP of each category. The top section shows the result with the setting of FED in \cite{Shakarami2021}. The '*' represents that AdamW is utilized as the optimizer with 12 epochs. The setting of mAP is $IoU=0.4$.}
    \label{tab:comparison_FED}
\end{table*}

\paragraph{Comparison with Deformable DETR.}
Deformable DETR \cite{Zhu2021DeformableDD} is a recent proposed detector which introduces transformer with deformable attention to object detection of small target detection problem. Although it can achieve surprising results on the BCCD dataset with higher mAP(The evaluation metrics of $AP_{50}$), it suffers convergence with more epochs and need dense operations per epoch. The result is shown in Table \ref{tab:comparison_previous}. The proposed method of TE-YOLOF converge much faster ($ 4 \times$) compared with Deformable DETR, and also achieve a higher mAP.

\paragraph{Comparison with YOLOv3.}
YOLOv3 is firstly used to blood cell detection in \cite{Zhang2019}. It achieve the goal of 86.79 in mAP, while we achieve 86.7 on the batchsize of 8 with the same setting in our device. While YOLOF \cite{chen2021you} can perform better in the same setting with AdamW\cite{Loshchilov2019DecoupledWD}. Table \ref{tab:comparison_previous} shows that TE-YOLOF can achieve better performance with lower epochs, the number of parameters and operations.

\paragraph{Comparison with FED.}
FED \cite{Shakarami2021} is the most efficient model with the minimal parameters in BCCD dataset. It utilize EfficientNet-B3 as backbone and YOLOv3 head as detection head. Modified the component based on YOLOv3
to observe better performance. Through the comparison of Table \ref{tab:comparison_FED}, compared to FED, TE-YOLOF-B0 can achieve better performance in mAP with lower parameters. When using the same backbone, TE-YOLOF can achieve mAP to 91.9 with the main contribution of increasing accuracy in RBC.

\subsection{Ablation Experiments}

\begin{table*}[t]
    \centering
    \begin{tabular}{ccccc}
        \toprule
        Backbone                        & EfficientNet-B0 & EfficientNet-B1 & EfficientNet-B2 & EfficientNet-B3 \\
        \midrule
        -                               & 87.4 & 88.3 & 88.4 & 89.0 \\
        + Depthwise Separable Module    & 88.7 & 88.8 & 89.1 & 89.2 \\
        \hline
        + MetaAconC                     & 88.7 & 88.9 & 89.1 & 88.9 \\
        + Swish                         & 89.6 & 89.5 & 89.4 & 90.0 \\
        + Mish                          & 89.8 & 90.2 & 89.9 & 90.5 \\
        \bottomrule
        \\
    \end{tabular}
    \caption{Comparison with different components. All the value is using the mAP in COCO detection evaluation metrics. '-' represents that we only replace the backbone with different EfficientNet version from original ResNet-50. '+Depthwise Separable Module' represents that using Depthwise Separable Convolution update Standard Convolution in decoder. The value of different activation function using in the model is displayed in the bottom section. }
    \label{tab:comparison_improvement}
\end{table*}

We run a number of ablations to analyze TE-YOLOF. We first provide an overall analysis of the three proposed components. Then we show the ablation experiments on detailed designs of each component. Results are shown in Table \ref{tab:comparison_improvement} and discussed in detail next. For the credibility, all experimental data are the average of 3 experiments.

\paragraph{Backbone of EfficientNet:}
Table \ref{tab:comparison_improvement} shows that EfficientNet as backbone is better than original backbone of ResNet-50 that result shows in Table \ref{tab:comparison_previous} with 87.6 mAP. The mAP has been improved according to the improvement of the EfficientNet version. EfficientNet-B0 has the lowest mAP with 87.8, while EfficientNet-B3 achieve the mAP to 89.0. The result of different backbone shows that the improvement of feature extraction capabilities is positive correlation with EfficientNet version.

\paragraph{Depthwise Separable Module:}
TE-YOLOF use Depthwise Separable Convolution replace the Standard Convolution in decoder. Figure \ref{fig:depthwise_separable_convolution} shows the architecture of this module. Depthwise Convolution can extract the features separately, while pointwise Convolution is able to fuse the data of each point on each feature map. The blood cell detection belongs to dense small target detection, so the ability of Depthwise Separable Convolution can show strength on this model. According to the Table \ref{tab:comparison_improvement}, the performance is increased by 0.2 to 1.3 points due to the contribution of Depthwise Separable Convolution in TE-YOLOF with different backbone.

\paragraph{Mish Activation function:}
The default activation is ReLU. Compared with three different activation functions, we can prove the effectiveness of the Mish activation function. MetaAconC \cite{ma2021activate} utilize $1 \times 1$ convolution to accomplish activation of the neurons or not. Table \ref{tab:comparison_improvement} shows that MetaAconC is unstable to the improvement in different backbone, while Swish and Mish is stable on the effectiveness of result. Mish is better than Swish in terms of the contribution to the result.

\subsection{Visual results}

\begin{figure*}[t]
    \centering
    \includegraphics[width=1\textwidth]{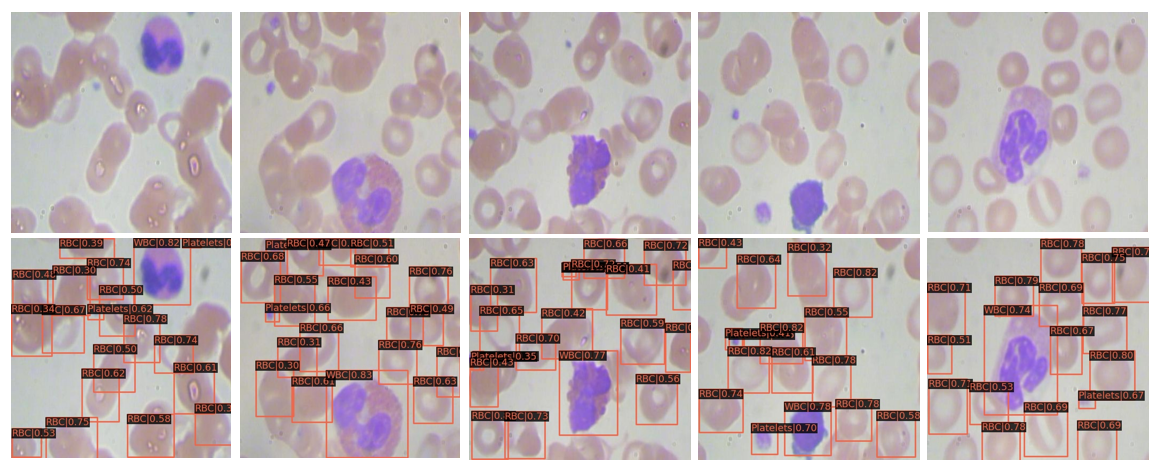}
    \caption{Visualization detection results on test set. Top to bottom: input images, detection results.}
    \label{fig:visual_result}
\end{figure*}

The detection results of the proposed model have been visualized in Figure \ref{fig:visual_result}. The model used is TE-YOLOF-B0 with the smallest amount of parameters and operations. It can be seen in the Figure \ref{fig:visual_result} that the model can detect the all the categories well. Platelets, RBC and WBC correspond to small, medium and large targets respectively. Our model can accurately identify different size categories, and can detect dense RBCs with high precision. 

\section{Discussion}
An traditional automated count of blood cells is usually performed via Flow Cytometry Instrumentation \cite{Bscher2019FlowCI}. For medical diagnostics in real situation, blood cell detection combined with computer vision technology can solve the challenge of detection speed and accuracy \cite{Li2019DeepCD}. One of the difficulties in blood cell detection is that the sparsity of platelets and WBCs and the denseness of RBCs cause the detection of various categories to be unbalanced. the FED model in \cite{Shakarami2021} can accurately detect platelets and WBCs, but the detection accuracy of RBCs can only reach 80.4 due to the density of RBCs. 

In this research, the proposed model has the ability to utilize Efficientnet's excellent feature extraction capabilities. With the help of the development of transfer learning theory in the field of computer vision, the proposed detector actually do not need time-consuming re-training. The pre-trained weights on the large dataset can be migrated to the new model, and then combined with small number of epochs, the final result will also meet expectations. 

In addition, The advantage of enlarging the features' receptive field in dilated convolution \cite{Yu2016MultiScaleCA} can be utilized to generate multi-scale features with different expansion factor. Stacking dilated convolution one by one without weight sharing can achieve the feature fusion effect of FPN. With its advantages, our model can achieve the balance and accuracy of the results in different size categories and the sparsity of the object. Conclusively, the proposed model, TE-YOLOF, overcomes the problem of low RBCs accuracy in FED as shown in Table \ref{tab:comparison_FED}. Thanks to the solution of the RBCs detection problem, the final result is significantly improved.

\section{Conclusion}
In this work, TE-YOLOF has been proposed to address blood cell detection problem. Based on YOLOF's excellent characteristics in dense object detection, compound scaling method, Depthwise Separable Convolution, and Mish activation function are utilized to lightweight transformation with better performance. The results of experiments and comparisons demonstrate that TE-YOLOF is more efficient than other existing methods for blood cell detection with minimum parameters. Besides, TE-YOLOF is a flexible detector because of the different backbone version that can be chosen to balance the precision and parameters. 

Due to the pandemic of COVID-19 coronavirus and late diagnosing dangerous and considering the effect of the blood cell detection in medical clinical diagnosis, it can be vital to develop an actual equipment using the proposed method. Besides, This solution can be considered to transfer to other medical detection fields as future work.

\section*{Acknowledgments}
This research was supported in part by Southwest Minzu University for Nationalities Excellent Student Cultivation Project (2021NYYXS78).

\bibliographystyle{unsrt}  
\bibliography{references}  
\end{multicols}

\end{document}